\documentclass{article}
\usepackage[preprint]{corl_2026} % Uncomment for pre-prints (e.g., arxiv); This is like ``final'', but will remove the CORL footnote.

\usepackage{graphicx}

\usepackage{algorithm}
\usepackage{algpseudocode}

\usepackage{amsmath}
\usepackage{bm}
\usepackage{mathtools}
\usepackage{amssymb}
\usepackage{amsthm}
\usepackage{amssymb}
\usepackage{xcolor}

\title{
ROVE: Unlocking Human Interventions for Humanoid Manipulation via Reinforcement Learning
}

\author{
{\bfseries Wei Xiao$^{1,2,*}$ \quad
Weiliang Tang$^{1,3,*}$ \quad
Yuying Ge$^{1,\dagger}$ \quad
Hui Zhou$^{1}$} \\
{\bfseries Yao Mu$^{4}$ \quad
Li Zhang$^{2}$ \quad
Yixiao Ge$^{1}$} \\[1mm]
{\normalfont\small $^{1}$XPENG Robotics \quad
$^{2}$Fudan University \quad
$^{3}$The Chinese University of Hong Kong} \\
{\normalfont\small $^{4}$Shanghai Jiao Tong University} \\
{\normalfont\small $^{*}$Equal contribution \quad $^{\dagger}$Corresponding author} \\
{\normalfont\small \href{https://xpeng-robotics.github.io/rove}{https://xpeng-robotics.github.io/rove}}
}

\begin{document}
	\maketitle
    
	\begin{figure}[ht]
        \vspace{-15pt}
		\centering
		\includegraphics[width=1.0\linewidth]{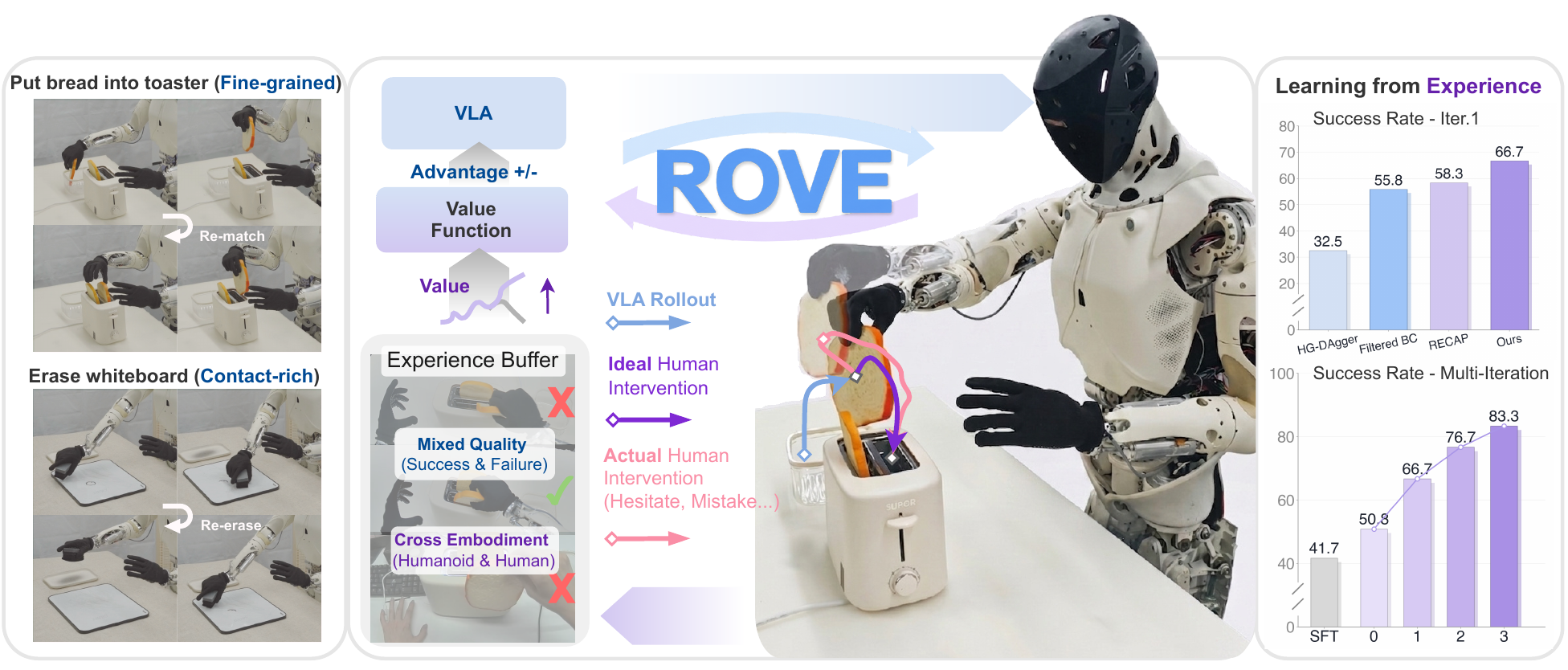}
        \vspace{-15pt}
\caption{
\textbf{ROVE learns from imperfect humanoid interventions.}
(Left) Our method recovers task progress through re-matching and re-erasing.
(Middle) Human interventions after near-failure VLA rollouts can deviate from ideal corrections in humanoid manipulation; our method learns a value function from mixed-quality and cross-embodiment experience to extract a better VLA policy.
(Right) It outperforms all baselines and consistently improves across multiple iterations.
}
        \vspace{-5pt}
		%\caption{An overview illustration for our submission.}
		\label{fig:teaser}
	\end{figure}

	%===============================================================================

	\begin{abstract}
		Human interventions provide crucial corrective signals for post-training Vision-Language-Action (VLA) models. However, enabling seamless humanoid interventions is a formidable systems challenge due to complex whole-body kinematics and dexterous-hand control. Consequently, the collected intervention trajectories are often suboptimal, and methods that rely on human interventions as expert supervision can absorb hesitant, inefficient, or even erroneous behaviors.
To address both the system and algorithmic challenges, we propose ROVE, a reinforcement learning framework for humanoid VLA post-training with imperfect human interventions. First, ROVE introduces a human-in-the-loop pipeline capable of collecting deployment and intervention data for humanoid manipulation. Second, it utilizes Optimistic Value Estimation (OVE) to prioritize high-value behaviors from mixed-quality trajectories. To further robustify value estimation, we incorporate cross-embodiment human experience videos to provide rich supervision for long-tailed failure and recovery modes. The resulting critic yields informative advantage signals, steering the VLA actor to focus on high-value behaviors rather than indiscriminately imitating all actions. On challenging real-world contact-rich and fine-grained humanoid manipulation tasks, ROVE outperforms experience-learning baselines and consistently improves across multiple rollout-intervention iterations.

	\end{abstract}

	% Two or three meaningful keywords should be added here
	\keywords{Vision-Language-Action Model, Reinforcement Learning, Humanoid Manipulation}

	%===============================================================================
    \section{Introduction}
Vision-Language-Action (VLA) models provide a promising foundation for general-purpose robot policies by grounding language instructions and visual observations into actions~\citep{bjorck2025gr00t, intelligence2025pi_05}.
Most recent progress has been largely demonstrated on robot-arm manipulation with parallel-jaw grippers, while extending VLA to humanoid robots with dexterous hands substantially increases the complexity of the robot dynamics and kinematics. 
This inherent complexity makes the humanoid robot fragile to accumulated errors stemming from whole-body poses, systematic errors, object contacts, etc. % \wl{This inherit complexity makes the humanoid error-pruning and fragile to accumulated errors stemming from whole-body poses, systematic errors, object contacts, etc.}This complexity makes the deployment much more sensitive to small errors in body and hand pose, object contact, or robot configuration.
As a result, VLA policies trained only on offline demonstrations often fail to reach satisfactory performance during deployment, struggling to handle deployment-time distribution shift. %\wl{fails to reach satisfying performance during deployment, struggling to handle deployment-time distribution shift} under deployment-time distribution shift and struggle to recover from out-of-distribution states.
Consequently, we seek a post-training solution to harness the deployment data for policy improvement. %\wl{Consequently, we seek for a post-training solution to harness the deployment data for policy improvement.}A natural next step is to turn the deployment itself into a source of improvement.
Reinforcement learning (RL)~\citep{intelligence2025pi} and interactive imitation learning (IIL)~\citep{ross2011reduction, kelly2019hg} allow robots to collect autonomous rollouts, observe their own failures, and incorporate human corrections. 
Recent post-training methods for VLA have shown that the deployment experience can improve pretrained policies beyond the offline demonstration tuning. 
Yet extending this paradigm to humanoid manipulation introduces a new difficulty: human teleoperation itself can be suboptimal, and this issue becomes especially severe during intervention takeover.

In a robot-arm with grippers setup, teleoperation interfaces such as 3D space mouse or leader-follower arms can often provide smooth and effective corrective actions~\citep{luo2025precise, zhao2023learning}. 
In contrast, humanoid teleoperation with dexterous hands is harder to master due to difficulty in whole-body coordination, dexterous manipulation, and joint-limit handling under imperfect retargeting and limited force or tactile feedback. %\wl{is less accurate due to difficulty in whole-body coordination, lack of the active tactile feedback for dexterous manipulation, and imperfect retargeting} because the robot dynamics and kinematics are substantially more complex: operators must coordinate body motion, dexterous-hand actions, contact timing, and joint-limit constraints under imperfect retargeting and limited force or tactile feedback.
During the takeover, even skilled operators may hesitate, retract their hands, or spend a period of time adapting to the humanoid's current configuration.
Prior human-in-the-loop RL (e.g., HIL-SERL~\citep{luo2025precise}, RECAP~\citep{intelligence2025pi}) and IIL (e.g., HG-DAgger~\citep{kelly2019hg}) methods typically assume that human interventions provide optimal corrective actions, so intervention data can be directly treated as expert supervision or positive improvement signals. This assumption becomes fragile in humanoid manipulation: due to the large teleoperation gap, the takeover phase often contains hesitant, redundant, or retargeting-induced actions. The key challenge is therefore how to learn from suboptimal data.

In this work, we propose \textbf{ROVE}, an \textbf{R}L framework with \textbf{O}ptimistic \textbf{V}alue \textbf{E}stimation for humanoid VLA post-training.
ROVE builds a human-in-the-loop data collection pipeline that supports whole-body and dexterous-hand intervention during VLA rollouts. 
To account for suboptimal collected data, we decompose intervention episodes into rollout, adaptation, and recovery stages, and construct value labels with a conservative boundary at the end of adaptation. 
We further learn a state value function with \emph{Optimistic Value Estimation (OVE)}, which combines TD bootstrapping with expectile regression to estimate high-value recoverable behavior from mixed-quality data. 
%Because the critic predicts state values (V-function) rather than state-action values (Q-function)\wl{For V and Q $<-$ Do we need to assume the VLA reader's background not familiar with RL ?}
Beyond robot trajectories, the critic also learns from videos of humans, which provide cross-embodiment supervision for progress and recovery without requiring robot-aligned actions.
The learned critic provides advantage labels for advantage-conditioned policy extraction, allowing the VLA actor to emphasize high-value actions instead of imitating all collected actions uniformly.
As shown in Fig.~\ref{fig:teaser}, ROVE improves upon SFT and experience-learning baselines, including RL and IIL methods, and continues to gain performance over multiple rollout-intervention iterations.

\begin{figure}[t]
    \centering
    \includegraphics[width=1.0\linewidth]{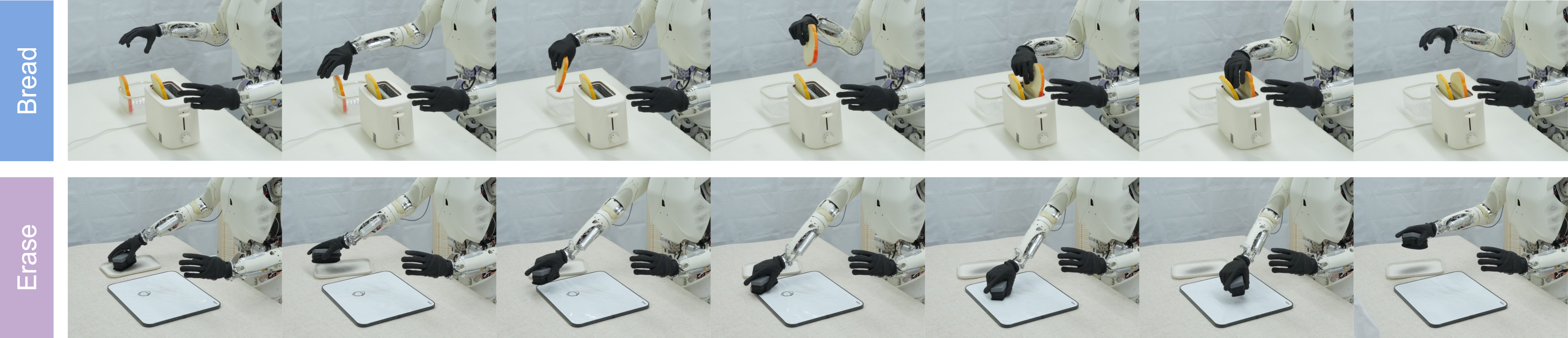}
    \vspace{-15pt}
    \caption{Task procedures for \textit{Put the bread into the toaster} (top) and \textit{Erase the whiteboard} (bottom).}
    \label{fig:task_protocol}
\end{figure}

Our main contributions are summarized as follows:
\textbf{(i)}
\textbf{Human-in-the-loop data collection pipeline for humanoid manipulation.}
    We build a whole-body human-in-the-loop collection pipeline and model the suboptimality of humanoid teleoperation, especially the noisy adaptation phase during intervention takeover;
\textbf{(ii)}
\textbf{Optimistic value estimation for heterogeneous experience.}
    We propose a state-value learning recipe that combines robot rollouts, human intervention trajectories, and human experience videos, producing robust advantage signals for policy extraction;
\textbf{(iii)}
\textbf{Iterative VLA policy improvement in real-world tasks.}
    Real-world experiments on two manipulation tasks, illustrated in Fig.~\ref{fig:task_protocol}, show that ROVE improves VLA policies from demonstrations, experience, and multiple rounds of rollout-intervention data.
% \end{itemize}

\section{Preliminary}

\textbf{Problem setting and notation.}
We formulate the humanoid manipulation problem as a Markov Decision Process (MDP) $\mathcal{M}=(\mathcal{S}, \mathcal{A}, \mathcal{P}, r, \gamma)$ in a standard RL setting~\citep{sutton1998reinforcement}, where $\mathcal{S}$ is the environment state space, $\mathcal{O}$ is the image observation space from humanoid's ego-view, $\mathcal{L}$ is the language space of the robot's task instruction. $\mathcal{A}$ is the action space of the robot, \mbox{$\mathcal{P}: \mathcal{S} \times \mathcal{A} \rightarrow \Delta(\mathcal{S})$} is the stochastic dynamics of the environment, \mbox{$r: \mathcal{S} \times \mathcal{A} \rightarrow \mathbb{R}$} is the reward function, the discount factor $\gamma \in [0,1]$.
ROVE starts from a generalist VLA model as the initial policy $a_{t:t+H-1} \sim \pi(\cdot|s_{t})$, $s_{t}=(o_{t}, l)$, $o_t \in \mathcal{O}, l \in \mathcal{L}$, $H$ is the action chunk horizon. We distinguish the environment state $x_t\in\mathcal{S}$ from the critic input $s_t=(o_t,l)$. The actor additionally receives proprioceptive state, omitted from policy notation for brevity. Given the robot policy and environment's transition $x_{t+1:t+H} \sim P(\cdot|x_{t}, a_{t:t+H-1})$, a set of trajectories is collected: \mbox{$\xi = (s_0, a_0, r_0, s_1, \cdots, a_{T-1}, r_{T-1}, s_T, r_T)$} through interactions between the policy $\pi$ and environment, and the distribution of trajectories is $\rho_{\pi}(\xi)$. We can define the discounted cumulative reward, or return, as $R(\xi) = \sum_{t=0}^T \gamma^{t} r_{t}$. The goal of RL is to maximize the expected discounted return: $\mathcal{J}(\pi) = \mathbb{E}_{\xi \sim \rho_{\pi}}[R(\xi)] = \mathbb{E}_{\xi \sim \rho_{\pi}}[\sum_{t=0}^T \gamma^{t} r_t]$.

\textbf{Value estimation.}
\label{pre:value_estimation}
For a policy $\pi$, the value function $V^\pi(s_t)$ is the expected future return $\mathbb{E}_{\rho_{\pi}(\xi)}[\sum_{i=t}^{T} \gamma^{i-t} r_{i} \mid s_t]$.
Then, we can define the advantage of the action chunk $a_{t:t+H-1}$ at state $s_t$ with $t+H\leq T$ as
$A^\pi(s_t,a_{t:t+H-1}) = \mathbb{E}_{\rho_{\pi}(\xi)} [\sum_{i=t}^{t+H-1} \gamma^{i-t} r_i + \gamma^H V^\pi(s_{t+H}) \mid s_t, a_{t:t+H-1}] - V^\pi(s_t)$.
In practice, we learn a parametric critic $V_{\phi}(s_t)$ from collected trajectories to estimate future return.
Monte-Carlo method (MC) regresses the critic toward the empirical discounted return, while Temporal-difference method (TD) bootstraps after the action chunk:
\begin{equation}
\begin{aligned}
\text{MC:}\quad
V_{\phi}(s_t) &\leftarrow \sum_{i=t}^{T} \gamma^{i-t} r_i,
&\quad
\text{TD:}\quad
V_{\phi}(s_t) &\leftarrow \sum_{i=t}^{t+H-1} \gamma^{i-t} r_i + \gamma^H V_{\bar{\phi}}(s_{t+H}).
\end{aligned}
\end{equation}
where $V_{\bar{\phi}}$ denotes a target network or a stop-gradient copy of $V_{\phi}$.
MC estimates are unbiased when complete returns are observed but can have high variance, while TD estimates reduce variance at the cost of bootstrapping bias.

    \begin{figure}[t]
    \centering
    \includegraphics[width=1.0\linewidth]{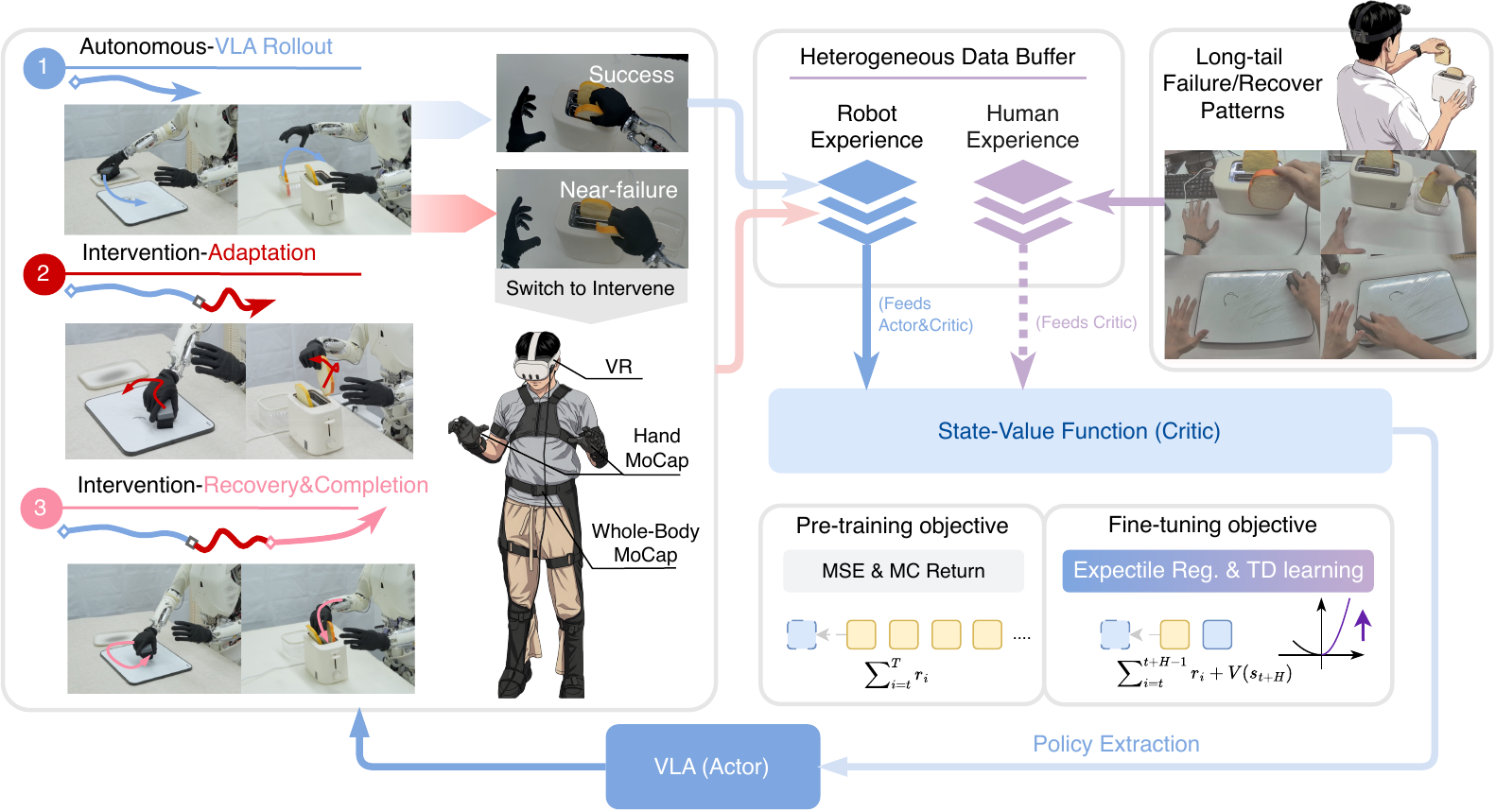}
    \vspace{-15pt}
    \caption{\textbf{Overview of ROVE framework.} A VLA actor collects autonomous rollouts and triggers whole-body human intervention near failure. The resulting trajectories are decomposed into rollout, adaptation, and recovery stages, and combined with cross-embodiment human experience to train a state-value critic. The critic provides advantage conditions for policy extraction, guiding the actor toward high-value recovery behaviors from mixed-quality data.}
    \label{fig:framework}
    \vspace{-10pt}
\end{figure}

\section{Methodology}
Fig.~\ref{fig:framework} summarizes the overall ROVE framework. The method first collects mixed-quality real-world experience on the humanoid robot, then learns a state-value critic from robot trajectories and human experience videos, and finally uses the resulting advantage signals to extract a stronger VLA policy. We describe these components in order.

\subsection{Human-in-the-loop data collection}

\textbf{Pipeline overview.}
We build a human-in-the-loop data collection pipeline that enables whole-body intervention during VLA policy rollouts. Each episode starts with autonomous execution, where the VLA policy predicts action chunks conditioned on the ego-view observation, task instruction, and proprioceptive state. When the supervisor detects a potential failure, the VLA action publisher is paused. Meanwhile, the motion-capture operator observes the robot state through a VR headset and aligns their body and hand pose with the current humanoid and dexterous hand configuration as closely as possible.

Once the operator is ready, the supervisor switches the robot from autonomous execution to teleoperation. The teleoperation interface maps the operator's whole-body and hand motion to humanoid control commands. Since the human pose and robot state may not be perfectly aligned at the takeover moment, we apply a command filter in the controller to smooth the transition and avoid abrupt changes in robot joint angles. The pause interval between policy rollout and human takeover is removed during post-processing.
During takeover, the operator starts from the terminal state of the VLA rollout, which is typically close to failure, and provides corrective actions until the task is completed. To balance the data distribution and increase trajectory diversity, we also initialize data collection from different task progress stages rather than only from the beginning of each task~\citep{Uchendu2022JumpStartRL}.

\textbf{Teleoperation gap and intervention suboptimality.}
Human-in-the-loop episodes are split into three stages: (1) Autonomous VLA rollout, (2) Intervention--adaptation, and (3) Intervention--recovery and task completion.
Unlike prior human-in-the-loop settings on standard robot arms, whole-body humanoid intervention with dexterous hands introduces a much larger teleoperation gap. 
This gap arises from the higher complexity of the robot morphology, the lack of force and tactile feedback, imperfections in motion retargeting, operator skill differences, and robot joint limits. As a result, a seamless transition from policy rollout to human intervention is difficult to achieve in practice. 
Even for highly skilled motion-capture operators, the initial takeover phase often involves unstable or hesitant motion.
Common behaviors in this phase include retracting the hand, continuing to manipulate from a near-failure state, or spending a period of time adapting to the robot's current configuration. We refer to this transient phase as the \emph{adaptation stage}. 
After the operator significantly advances task progress, the trajectory enters the \emph{recovery and completion stage}, where the operator continues the task until the task is completed.

\textbf{Reward design.}
Following the stage decomposition above, we assign rewards according to both task outcome and intervention phase:

\begin{equation}
    r_t =
    \begin{cases}
    0, & t=T,\ \text{success}, \\
    C_{\mathrm{fail}}, & t=T,\ \text{autonomous rollout failure}, \\
    C_{\mathrm{fail}}, & t=t_r,\ \text{end of the adaptation stage}, \\
    -1, & \text{otherwise}.
    \end{cases}
\end{equation}

Here, $T$ denotes the terminal timestep of an episode, and $t_r$ marks the end of the initial adaptation stage, after which intervention begins to produce sustained task progress. We therefore assign the penalty at $t_r$ rather than at intervention onset; this conservative boundary avoids mixing adaptation-stage alignment behavior into recovery labels (see Appendix~\ref{sec:value_label_sensitivity}).
During task fine-tuning, we set $C_{\mathrm{fail}}=-500$. During pretraining, we set $C_{\mathrm{fail}}$ to the negative mean episode length of each task, which provides a task-normalized failure scale.
We set $\gamma=1$ in our experiments. Except at the adaptation boundary, non-terminal steps receive a penalty of $r_t=-1$, encouraging shorter completion times.

\subsection{Value function training}

\textbf{Data recipe and pretraining objective.}
The value function is trained in stages. We first pretrain the critic on large-scale robot and egocentric human demonstrations~\citep{Hoque2025EgoDexLD} to learn a general notion of task progress, then fine-tune it for the downstream task.
In later iterations, we train the critic on autonomous rollouts and intervention trajectories, with task-relevant human experience videos. These videos provide cross-embodiment examples of recovery and completion that are sparse in robot experience, improving value estimates on partial-progress and near-failure states.
For pretraining, we use a Monte-Carlo regression objective to fit the cumulative return from each state:
\begin{equation}
\mathcal{L}_{\mathrm{MC}}(\phi)
=
\mathbb{E}_{(s_t,\xi)\sim\mathcal{D}}
\left[
\left(
V_{\phi}(s_t)
-
\sum_{i=t}^{T} \gamma^{i-t} r_i
\right)^2
\right].
\end{equation}
This objective estimates the average return of the dataset behavior distribution. It is stable for large-scale pretraining on heterogeneous data from different tasks and embodiments, and provides a robust initialization for task-level fine-tuning.

\textbf{Optimistic value estimation.}
For fine-tuning, however, estimating only the average return can be overly conservative. Our training data contain many suboptimal teleoperated behaviors, failed autonomous rollouts, and transient adaptation segments after takeover. If the critic simply averages over these trajectories, the estimated value can be much lower than the value of the best recoverable behavior from the same state.

To address this issue, we use \emph{Optimistic Value Estimation (OVE)}, combining an $H$-step TD bootstrap~\citep{park2026horizon} with expectile regression. Related expectile-based value-learning objectives have been studied in~\citep{kostrikov2021offline, park2023hiql, xu2022policyguided}. Given a transition segment of length $H$, we define the bootstrapped target as
\begin{equation}
    \hat{V}_t
    =
    \sum_{i=t}^{t+H-1} \gamma^{i-t} r_i
    +
    \gamma^H V_{\bar{\phi}}(s_{t+H}),
\end{equation}
\begin{equation}
    \mathcal{L}_{\mathrm{OVE}}(\phi)
    =
    \mathbb{E}_{(s_t,\xi)\sim\mathcal{D}}
    \left[
    \left|
    \tau - \mathbf{1}\{\hat{V}_t - V_{\phi}(s_t) < 0\}
    \right|
    \left(\hat{V}_t - V_{\phi}(s_t)\right)^2
    \right],
\end{equation}
where $V_{\bar{\phi}}$ is an EMA target critic updated from $V_{\phi}$. Instead of fitting the bootstrapped target $\hat{V}_t$ with a symmetric squared loss, we use an expectile loss, where $\tau$ controls the degree of optimism. When $\tau=0.5$, the objective reduces to standard mean regression; as $\tau$ approaches $1$, the critic places more weight on higher-value outcomes. Thus, OVE estimates an in-distribution optimistic statistic: it favors better recoveries observed in the data without querying out-of-distribution actions.

\subsection{Policy improvement and implementation details}
The VLA actor is trained with advantage conditioning to indicate whether an action chunk is predicted to improve task progress. Algorithm~\ref{alg:rove_policy_improvement} summarizes the iterative loop.
In each iteration, we update critic and actor sequentially: train the critic with $\mathcal{D}^{critic}_k$, compute advantage statistics on $\mathcal{D}^{actor}_k$, assign binary advantage via a threshold, and fine-tune the actor (details in Appendix~\ref{pre:policy_extraction}).
Our experiments use the IRON-R01-1.11 humanoid robot with a $50$-dimensional proprioceptive state and action space covering body joints and dexterous hands.
With limited task-specific robot data, conditioning on the high-dimensional proprioceptive state can overfit to brittle joint-level cues. We therefore apply state dropout and perturbation during training to rely more on visual task progress (details in Appendix~\ref{sec:additional_training}).
    \section{Experiments}
Our experiments are conducted in two real-world humanoid manipulation tasks: a contact-rich task (\textit{Erase the whiteboard}) and a fine-grained task (\textit{Put the bread into the toaster}). The analysis is organized around two aspects: policy improvement and value estimation.

\begin{figure}[t]
    \centering
    \includegraphics[width=1.0\linewidth]{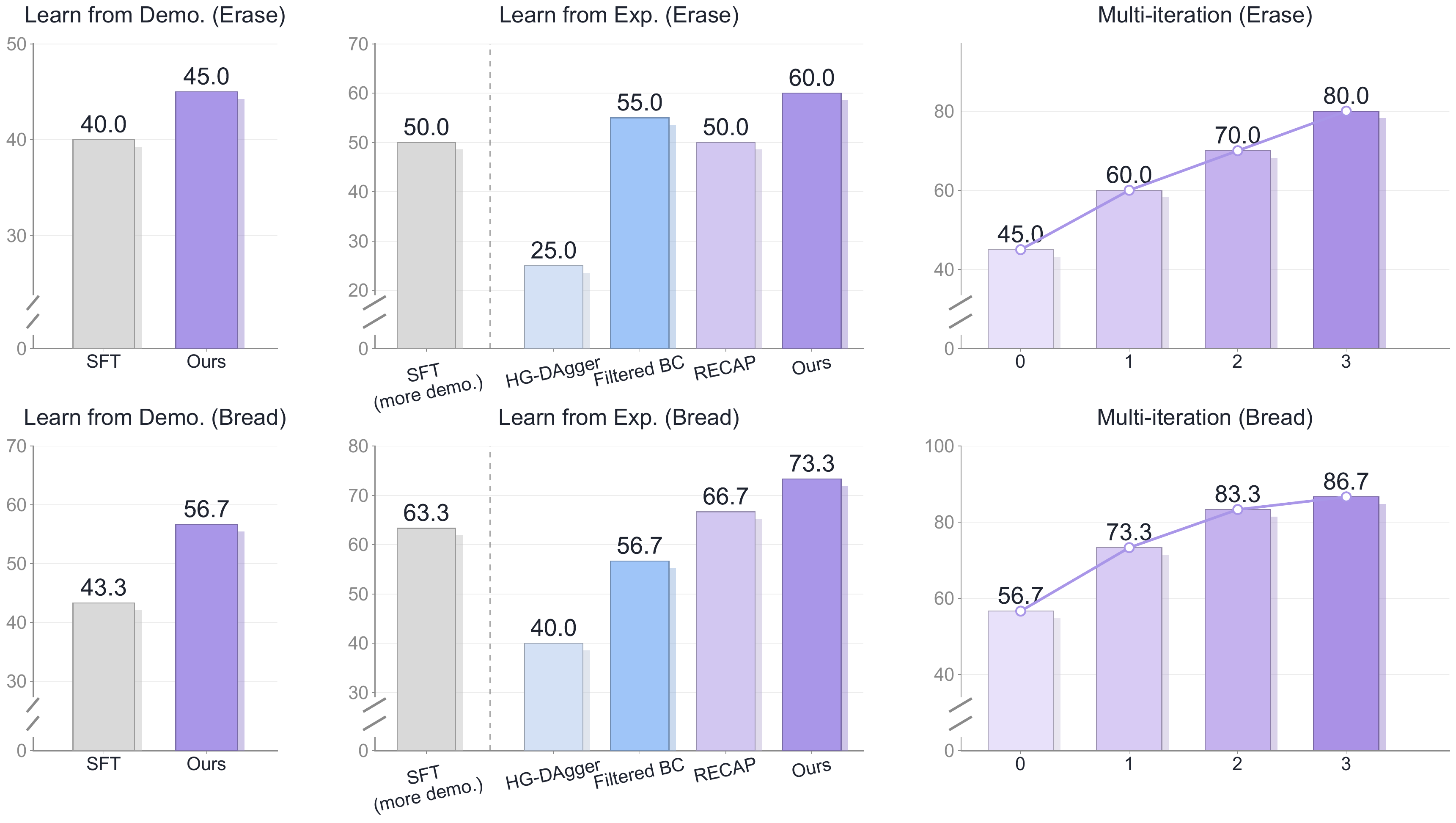}
    \vspace{-15pt}
    \caption{\textbf{Policy improvement results on two real-world humanoid manipulation tasks.}
    ROVE (Ours) outperforms SFT in the demonstration-only setting, achieves the best average performance among experience-learning methods, and consistently improves across multiple iterations of rollout and intervention data.}
    \vspace{-10pt}
    \label{fig:policy_improvement}
\end{figure}

\subsection{Policy improvement results and analysis}

\textbf{Learn from demonstrations.}
We fine-tune the pretrained VLA with either standard SFT or ROVE using teleoperated demonstrations. Fig.~\ref{fig:policy_improvement} shows that ROVE improves over SFT on both tasks. This suggests that humanoid teleoperation demonstrations can be suboptimal due to the teleoperation gap and operator variability, and that value-guided extraction recovers higher-quality behavior than uniform imitation.

\textbf{Learn from experience.}
We compare ROVE with experience-learning baselines, including HG-DAgger~\citep{kelly2019hg}, Filtered BC, and RECAP~\citep{intelligence2025pi} (details in Appendix~\ref{sec:baselines}).
Fig.~\ref{fig:policy_improvement} shows that ROVE achieves the best average success rate across tasks. A notable result is that HG-DAgger performs poorly, even below the base demonstration-only policy on both tasks, and its learned policy often exhibits hesitant behavior. This pattern reflects the suboptimality of directly imitating intervention data. Compared with RECAP, the remaining gap reflects the combined effect of critic quality and advantage assignment of post-adaptation intervention segments. Compared with Filtered BC, the gap suggests that RL-style policy learning provides additional gains beyond BC-style data filtering, particularly in how negative samples are incorporated during policy optimization.
As a control, scaling up SFT demonstrations yields additional gains, while the failure-recovery behaviors of ROVE are rarely observed in these demonstration-only policies at deployment.

\textbf{Iterative improvement.}
We collect a comparable amount of real-world rollout and intervention data in each iteration, then update the value function and policy from the previous iteration.
As shown in Fig.~\ref{fig:policy_improvement}, ROVE consistently improves across three iterations on both tasks. The success rate increases from 45.0\% to 80.0\% on \textit{Erase the whiteboard}, and from 56.7\% to 86.7\% on \textit{Put the bread into the toaster}. This demonstrates that it forms a closed-loop improvement process: better policies collect more informative experience, and the value function provides increasingly useful advantage signals for subsequent policy updates.

\subsection{Value estimation analysis}

We then examine the value function behind policy gains, focusing on how human experience and optimistic value estimation affect the reliability and resolution of value estimates.

\begin{figure}[t]
    \centering
    \includegraphics[width=1.0\linewidth]{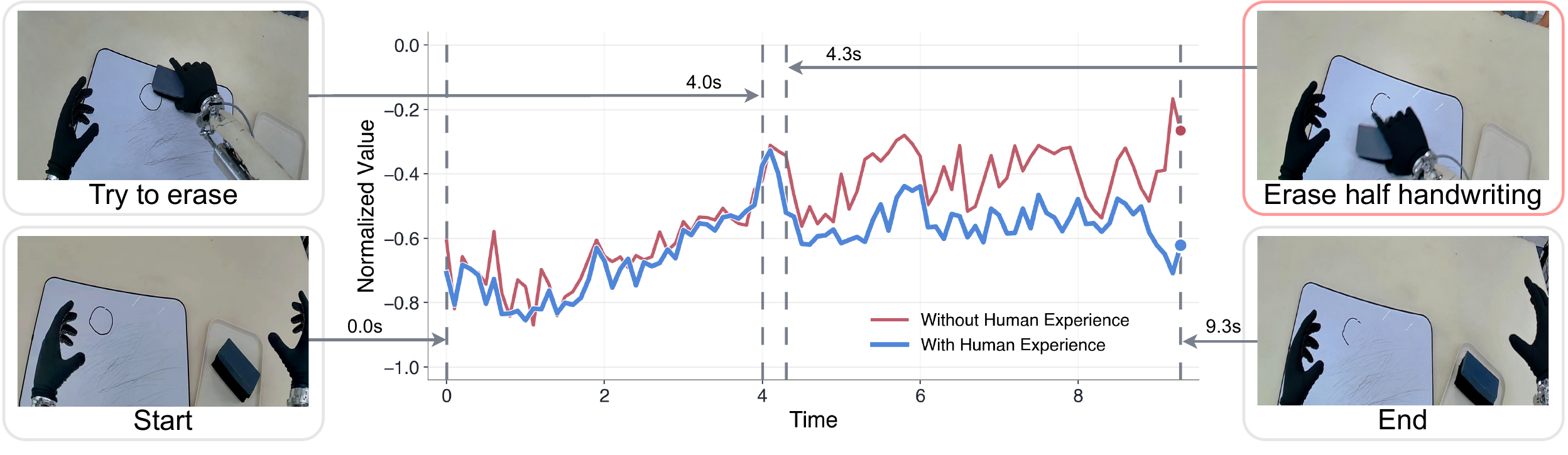}
    \vspace{-15pt}
    \caption{\textbf{Human experience improves value estimation.}
    Adding human experience helps the critic assign lower values to incomplete erasing states and better reflect true task progress.}
    \label{fig:value_ablation_human}
    \vspace{-10pt}
\end{figure}

\textbf{Human experience.}
Fig.~\ref{fig:value_ablation_human} compares two critics trained with and without human experience videos on the same held-out trajectory.
Without human experience, the critic tends to overestimate intermediate states where the robot only partially erases the handwriting. In contrast, the critic trained with human experience assigns lower values to these incomplete states and produces a value curve that better follows actual task progress.
Human trajectories provide examples of recovery from near-failure or incomplete states that are rarely covered by autonomous rollouts, yielding more reliable advantage estimates for policy improvement.

\begin{figure}[t]
    \centering
    \includegraphics[width=1.0\linewidth]{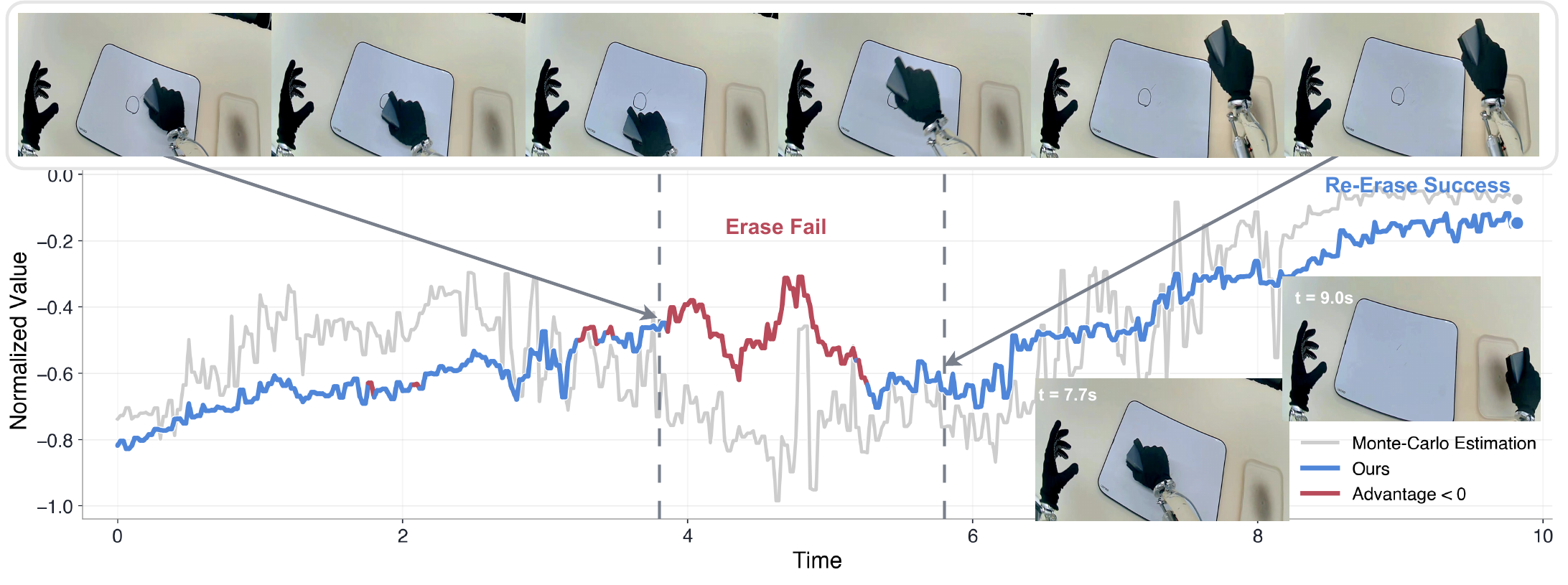}
    \vspace{-15pt}
    \caption{\textbf{OVE provides sharper value estimates than Monte-Carlo estimates}, producing clearer negative-advantage regions during failure and recovery.}
    \label{fig:value_ablation_ove}
    \vspace{-10pt}
\end{figure}

\textbf{Value learning objective.}
Fig.~\ref{fig:value_ablation_ove} compares OVE with MC value estimation on a held-out trajectory containing both failure and recovery segments.
The MC critic estimates the average return of the full data buffer and tends to produce noisy, less discriminative values when successful recoveries and failed rollouts are mixed. OVE instead produces a more structured value curve that better follows task progress.
In particular, during the erase-failure segment, OVE assigns lower values and yields negative-advantage regions, indicating that the critic recognizes actions that move the trajectory away from successful completion.
When the robot later re-erases the board and recovers the task, the value increases accordingly. These observations are consistent with OVE providing a sharper training signal for policy extraction by distinguishing harmful actions from recoverable progress while remaining in-distribution over observed trajectories.
    \section{Related work}
\label{sec:related_work}

\textbf{Reinforcement learning for VLA post-training.} Reinforcement learning has become a key route for improving VLA policies beyond imitation-only tuning. Prior work spans simulation-scale on-policy RL~\citep{li2025simplevla, lu2025vla}, real-world experience learning with advantage conditioning~\citep{intelligence2025pi}, world-model-based policy improvement~\citep{yang2026rise, zhang2025reinforcing, zhu2025wmpo, Li2025VLARFTVR}, and fleet-scale offline-to-online adaptation~\citep{wang2026learning}. Collectively, these results establish the importance of learning from deployment experience for VLA improvement. ROVE extends this paradigm to real-world humanoid manipulation with dexterous hands, where higher embodiment complexity makes experience distributions noisier and critic robustness more critical than in robot-arm settings.

\textbf{Human-in-the-loop robot learning.} Human-in-the-loop learning improves data efficiency by letting humans recover policy failures during deployment~\citep{liu2025robot, luo2025precise}. Recent work extends this idea to preference-style refinement~\citep{xia2026human}, dexterous interventions~\citep{han2026dexhil}, and seamless hand-arm takeover~\citep{li2026hand}, often under assumptions that intervention corrections can be treated as high-quality supervision. As embodiment complexity increases, however, takeover misalignment and adaptation behavior naturally become more prominent in intervention trajectories. ROVE is designed for this regime by treating intervention data as mixed-quality and learning value signals that separate recovery behavior from naturally occurring takeover noise.

A more comprehensive discussion is provided in Appendix~\ref{sec:related_work_extended}.
    \section{Conclusion}

We presented ROVE, an RL framework for post-training humanoid VLA policies from demonstrations, autonomous rollouts, human interventions, and human experience videos. The core challenge is that humanoid teleoperation data are not uniformly expert, especially during intervention. We address this with stage-aware intervention labeling, an optimistic state-value critic learned from heterogeneous experience, and advantage-conditioned actor training. Real-world experiments show that ROVE improves over SFT, compares favorably with experience-learning baselines, and continues to improve over rollout-intervention iterations. These results support ROVE as a practical path for robust humanoid VLA post-training from informative but imperfect deployment experience.

\section{Limitations}

First, human experience is currently used only for value learning, not for direct policy learning. Future work could use representation-level supervision to let policies benefit from human videos.
Second, our system lacks end-effector sensing such as wrist cameras and tactile feedback, which limits precise manipulation.
Third, we have not yet extended ROVE to loco-manipulation. Future extensions will need to handle locomotion and intervention safety.
Finally, ROVE is mainly an offline or iterative offline RL framework. Extending it to online RL would require efficient exploration and stable deployment-time updates.

	%===============================================================================

	%===============================================================================

	%===============================================================================

	%===============================================================================

	\clearpage
	% The acknowledgments are automatically included only in the final and preprint versions of the paper.
	% \acknowledgments{If a paper is accepted, the final camera-ready version will (and probably should) include acknowledgments. All acknowledgments go at the end of the paper, including thanks to reviewers who gave useful comments, to colleagues who contributed to the ideas, and to funding agencies and corporate sponsors that provided financial support.}
    %===============================================================================

	% no \bibliographystyle is required, since the corl style is automatically used.
	\bibliography{example} % .bib

    \clearpage
    \appendix
    \section{Extended related work}
\label{sec:related_work_extended}
This section provides an extended related-work discussion with additional representative directions and technical context.

\subsection{Reinforcement learning for VLA post-training}
Reinforcement learning has shown the potential to improve robot policies beyond imitation learning by optimizing task outcomes from physical interaction. Recent systems demonstrate strong performance on real-world manipulation with efficient off-policy RL~\citep{luo2024serl}, human corrections~\citep{luo2025precise}, and carefully designed training pipelines~\citep{lei2025rl}. In parallel, reinforcement post-training for Vision-Language-Action (VLA) models has become an active direction. \citet{li2025simplevla, lu2025vla} improve VLAs via on-policy RL~\citep{schulman2017proximal, shao2024deepseekmath} in simulation, while \citet{intelligence2025pi} shows experience learning with advantage conditioning in real-world deployment. \citet{yang2026rise, zhang2025reinforcing, zhu2025wmpo, Li2025VLARFTVR} improve VLAs through imagined rollouts and value-based evaluation in learned world models, and \citet{wang2026learning} studies fleet-scale offline-to-online RL for continual adaptation of generalist policies. Additional interfaces include representation-level residual RL~\citep{xu2026rl, johannink2019residual, gong2025robust} and latent-space RL with progress-based filtering~\citep{li2025gr, wagenmaker2025steering}. Within this landscape, ROVE targets real-world humanoid post-training with dexterous hands, where embodiment complexity makes experience statistics and critic robustness especially important.

\subsection{Human-in-the-loop robot learning}
Human-in-the-loop robot learning improves data efficiency by allowing humans to correct policy failures during deployment. \citet{liu2025robot} studies human-in-the-loop autonomy where assistance is used both for failure recovery and policy improvement over time. \citet{luo2025precise} integrates demonstrations and online corrections into real-world RL, showing substantial acceleration. Action preference optimization further uses human-assisted rollouts for preference-style refinement~\citep{xia2026human}. For dexterous settings, \citet{han2026dexhil} extends human-in-the-loop post-training to coordinated arm-hand interventions, while \citet{li2026hand} shows that takeover mismatch can cause abrupt gesture jumps and post-intervention degradation. Together, these studies highlight both the value of intervention data and the growing importance of takeover quality as embodiment complexity increases. ROVE follows this direction but treats intervention trajectories as inherently mixed-quality in high-complexity humanoid teleoperation, rather than assuming uniformly high-quality corrective supervision.

\subsection{Value learning for offline and off-policy data}
Value learning is central to extracting improved behavior from offline and off-policy data, especially when datasets mix successful recoveries with suboptimal teleoperation and failed rollouts~\citep{levine2020offline}. In classical offline RL, a state-action value function $Q(s,a)$ is often trained on static or heterogeneous data, leading to distributional shift and overestimation of out-of-distribution actions. \citet{kumar2020conservative} mitigates this issue by penalizing over-optimistic $Q$-values, while in-sample methods such as \citet{xu2023offline, kostrikov2021offline} avoid querying unseen actions through implicit regularization. Recent robot and VLA post-training systems also increasingly use value or critic models to improve policies from deployment data: \citet{lei2025rl, wang2026learning} combine offline and online learning with IQL~\citep{kostrikov2021offline} or its distributional variant in real-world settings. ROVE follows this broad value-learning motivation but learns a state value $V(s)$ rather than a state-action value $Q(s,a)$, allowing the critic to use robot rollouts, human intervention trajectories, and cross-embodiment human experience videos under a unified objective for optimistic state-value estimation.

\section{Sensitivity of value-label construction}
\label{sec:value_label_sensitivity}

\begin{table}[t]
\centering
\small
\caption{Sensitivity of value-label construction. We vary critic horizon and penalty reward timing $t_r$ while keeping all other settings fixed on the \textit{Erase the whiteboard} task.}
\label{tab:value_label_sensitivity}
\renewcommand{\arraystretch}{1.12}
\setlength{\tabcolsep}{5pt}
\begin{tabular}{lcc}
\hline
Setting & Success rate & Change vs.\ default \\
\hline
\textbf{ROVE (default)} ($H=16$, $t_r$ after adaptation) & \textbf{80\%} & - \\
Long critic horizon ($H=50$) & 65\% & -15\% \\
$t_r$ at intervention start ($H=50$) & 50\% & -30\% \\
\hline
\end{tabular}
\end{table}

Table~\ref{tab:value_label_sensitivity} studies two implementation choices in value-label construction: the horizon used for TD bootstrapping and advantage computation, and the timing $t_r$ of the penalty reward.
Increasing the horizon from $H=16$ to $H=50$ reduces success from 80\% to 65\%, indicating that an overly long horizon makes advantage labels less responsive to local recovery progress. Moving the penalty reward to takeover start further reduces success to 50\%, because adaptation-stage alignment behavior is then mixed into recovery labels. This result does not suggest that the boundary requires precise manual tuning; instead, it reveals an asymmetric risk in intervention data. A slightly delayed boundary is conservative and may omit some early recovery actions, whereas an overly early boundary introduces noisier adaptation-stage behavior into value targets. Since OVE preferentially propagates high-value continuations observed in data, this early-boundary noise can be amplified into misleadingly optimistic labels. We therefore place the boundary after the adaptation stage so OVE propagates value through cleaner recovery segments.

\section{Comparison with IQL}
\label{sec:iql_comparison}
IQL~\citep{kostrikov2021offline} is an offline RL method that learns an in-distribution optimistic value through expectile regression. In the standard IQL formulation, a state-action critic $Q_{\theta}(s_t,a_t)$ is first trained with a Bellman backup:
\[
\mathcal{L}_{Q}^{\mathrm{IQL}}(\theta)
=
\mathbb{E}_{(s_t,a_t,r_t,s_{t+1})\sim\mathcal{D}}
\left[
\left(Q_{\theta}(s_t,a_t)-r_t-\gamma V_{\psi}(s_{t+1})\right)^2
\right].
\]
The value function is then fitted as an upper expectile of the action-value distribution under dataset actions:
\[
\mathcal{L}_{V}^{\mathrm{IQL}}(\psi)
=
\mathbb{E}_{(s_t,a_t)\sim\mathcal{D}}
\left[
\left|
\tau - \mathbf{1}\{Q_{\theta}(s_t,a_t)-V_{\psi}(s_t)<0\}
\right|
\left(Q_{\theta}(s_t,a_t)-V_{\psi}(s_t)\right)^2
\right].
\]
The resulting advantage is computed as
\[
A(s_t,a_t)
=
Q_{\theta}(s_t,a_t)-V_{\psi}(s_t),
\]
and is then used for CFGRL-style policy extraction~\citep{frans2025diffusion}. This design avoids explicitly querying out-of-distribution actions, but it still relies on a state-action value function whose action input must be well defined within a single embodiment and action space.

We adopt the same principle of in-distribution optimism, but use it differently. Instead of learning $Q(s_t,a_t)$, we directly learn a state value $V_{\phi}(s_t)$ with a bootstrapped target:
$
\hat{V}_t
=
r_t
+\gamma V_{\bar{\phi}}(s_{t+1}).
$
For brevity, we present the one-step TD form here to align with the IQL notation above.
The expectile objective is applied between $\hat{V}_t$ and $V_{\phi}(s_t)$, yielding an optimistic state-value estimator. This choice is important for our setting because the data used for value learning include robot rollouts, human intervention trajectories, and cross-embodiment human experience videos. These sources do not always share a compatible robot action representation, making a unified $Q(s,a)$ difficult to define and prone to action-space overfitting. A state-value critic, in contrast, allows ROVE to use heterogeneous experience for value learning while still producing advantage labels for policy extraction.

\begin{figure}[t]
    \centering
    \includegraphics[width=1.0\linewidth]{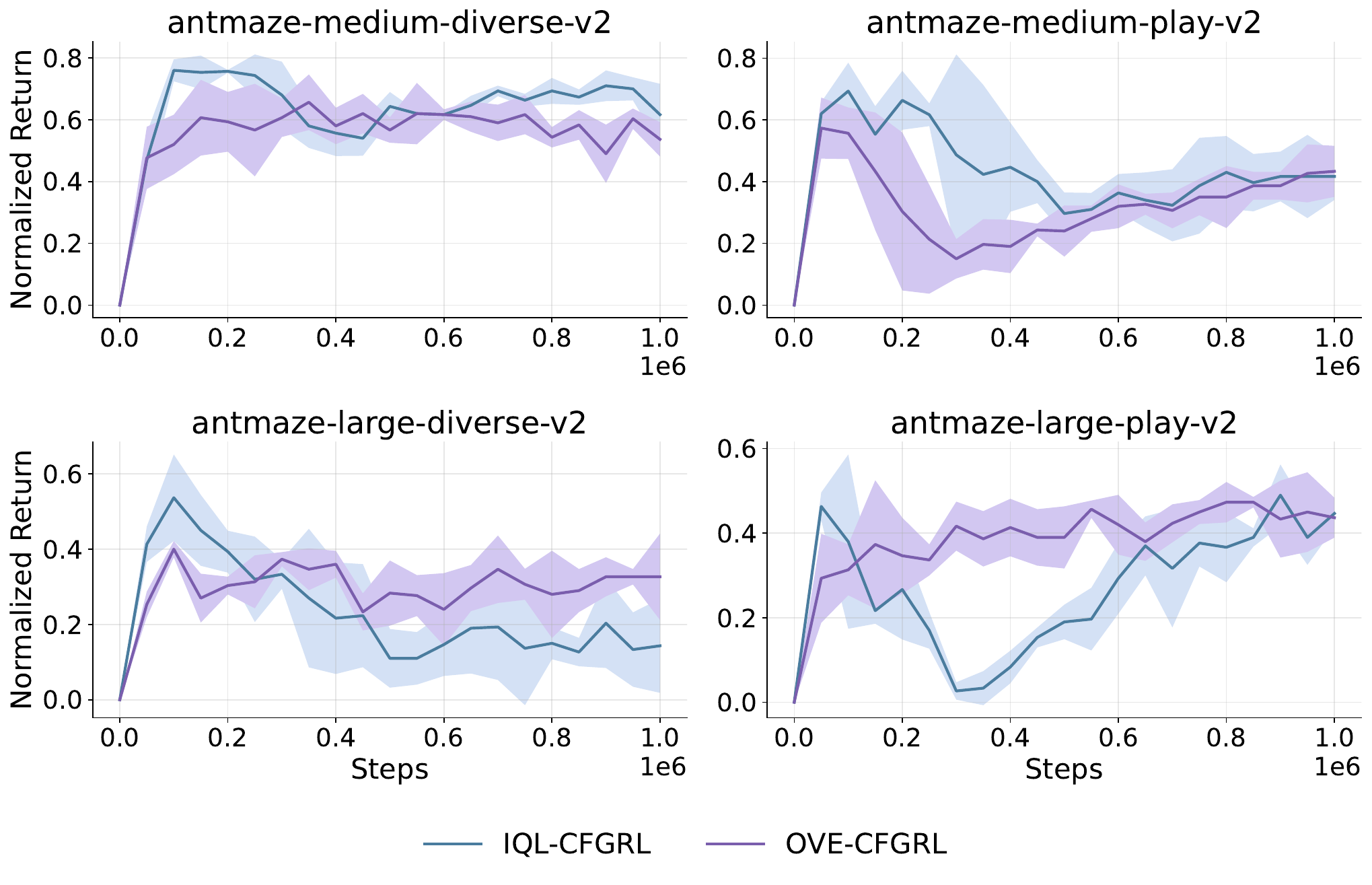}
    \caption{Performance comparison between OVE and IQL on D4RL-AntMaze tasks.}
    \label{fig:iql_comparison}
\end{figure}

We additionally compare OVE with IQL on D4RL-AntMaze tasks~\citep{fu2021d4rldatasetsdeepdatadriven} in Fig.~\ref{fig:iql_comparison}. The results show that OVE achieves competitive performance across medium and large maze variants, supporting that state-value optimistic estimation can serve as a practical alternative to Q-based expectile learning.

\section{Task description and evaluation protocol}
\label{sec:task_description}

Fig.~\ref{fig:task_protocol} illustrates the manipulation procedure for each task.
For \textit{Put the bread into the toaster}, the task begins with bread slices in a container and a two-slot toaster on the table. A trial proceeds through reaching into the container, grasping one slice, lifting and transporting it above the toaster slot, inserting the bread, releasing it, and retracting the hand. A trial is counted as successful if the bread is fully inserted into the toaster slot and remains stable.
For \textit{Erase the whiteboard}, the task begins with an eraser on a tray and handwriting marks on the board. A trial proceeds through picking up the eraser, moving to the marked region, establishing contact, wiping across the assigned marks with sufficient coverage, and lifting the eraser after erasing. A trial is successful if all handwriting marks assigned to that trial are removed.
To complete these tasks, a humanoid robot requires coordinated dexterous hand manipulation, head motion(e.g., locating handwriting on the board), and waist adjustment (left-right rotation or forward lean) to maintain effective contact and reachability.

\begin{figure}[t]
    \centering
    \includegraphics[width=1.0\linewidth]{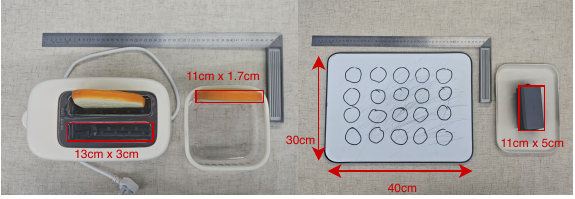}
    \caption{Evaluation objects and layout for our tasks.}
    \label{fig:eval_protocol}
\end{figure}

Fig.~\ref{fig:eval_protocol} summarizes the evaluation objects and scene layout used in our experiments.
For \textit{Erase the whiteboard}, we draw twenty handwriting marks uniformly distributed on the whiteboard and run one evaluation trial per mark, for twenty trials in total. For \textit{Put the bread into the toaster}, we run thirty independent trials per policy. Success rate is computed as the number of successful trials divided by the total number of trials.

\section{Baselines}
\label{sec:baselines}

We compare ROVE against three experience-learning baselines, implemented on the same humanoid platform, data collection protocol, and VLA backbone.

\textbf{HG-DAgger}~\citep{kelly2019hg,ross2011reduction} follows the interactive imitation learning paradigm: during policy rollouts, a human operator takes over when the policy approaches failure, and the corrected trajectories are aggregated into the training set for behavior cloning. In our implementation, HG-DAgger uses the same rollout--intervention episodes as ours but treats all human-intervention actions as positive supervision.

\textbf{RECAP}~\citep{intelligence2025pi} follows the $\pi^*_{0.6}$-style experience-learning recipe: a critic provides progress or advantage signals from deployment experience, and the VLA actor is fine-tuned with advantage-conditioned behavior cloning to prefer high-value action chunks over uniform imitation. For a controlled comparison with ROVE, RECAP uses the same robot data and the same reward-label construction pipeline; the key differences are that its critic training objective differs from OVE and its labeling strategy treats human intervention segments after the adaptation stage as uniformly positive advantages.

\textbf{Filtered BC} filters rollout and intervention data before behavior cloning. Following progress-based data refinement ideas~\citep{li2025gr}, we retain trajectory segments whose critic-estimated advantage exceeds a threshold and discard segments associated with failed or low-advantage states. To isolate the policy-learning objective, Filtered BC uses the same critic checkpoint as ours for advantage scoring and performs thresholded data filtering plus standard BC.

\section{Policy extraction via advantage conditioning}
\label{pre:policy_extraction}
Advantage conditioning provides a supervised way to extract an improved policy from experience~\citep{frans2025diffusion, intelligence2025pi}. Let $I$ denote an improvement event indicating that an action chunk is preferred over the typical behavior of a reference policy $\pi_{\mathrm{ref}}$ at the same state. Given an advantage estimate $A^{\pi_{\mathrm{ref}}}(s_t,a_{t:t+H-1})$, policy improvement can be written as a target action distribution that reweights the reference policy by the likelihood of improvement:
\begin{equation}
\hat{\pi}(a_{t:t+H-1}|s_t) \propto \pi_{\mathrm{ref}}(a_{t:t+H-1}|s_t)\,p(I|A^{\pi_{\mathrm{ref}}}(s_t,a_{t:t+H-1}))^\beta .
\end{equation}
Since the improvement event is determined by the advantage label, $p(I|A^{\pi_{\mathrm{ref}}})$ can be implemented as an additional condition on the action model. Equivalently, we train a conditional policy $\pi_{\theta}(a_{t:t+H-1}|s_t,I)$ by behavior cloning on trajectories annotated with discretized or binary advantage labels:
\begin{equation}
\max_{\theta}\; \mathbb{E}_{(s_t,a_{t:t+H-1},I)\sim\mathcal{D}}\left[\log \pi_{\theta}(a_{t:t+H-1}|s_t,I)\right].
\end{equation}
In implementation, we instantiate the improvement event $I$ as a binary advantage label computed by the learned critic. Before actor training, we set the threshold $\eta_k$ to the $70$th percentile of the advantage distribution. Each training action chunk is then labeled as positive if its advantage exceeds $\eta_k$ and negative otherwise. For CFGRL-style conditioning~\citep{frans2025diffusion}, we append a text condition to the instruction and train the VLA actor with the standard action prediction loss. At inference time, the actor uses classifier-free-guidance-style decoding: it predicts action velocities under both positive and negative conditions, and combines them as $v_{\mathrm{cfg}} = v_{\mathrm{neg}} + \beta (v_{\mathrm{pos}} - v_{\mathrm{neg}})$. This guides generation toward high-advantage actions rather than simply querying a single positive condition.

\section{Additional training and implementation details}
\label{sec:additional_training}
% Alg.~\ref{alg:rove_policy_improvement} uses \texttt{algpseudocode} commands (\verb|\Require|, \verb|\For|, \verb|\State|). Load \texttt{\string\usepackage\{algpseudocode\}} in the preamble (and do not also load the legacy \texttt{algorithmic} package).

This appendix records the full policy-improvement procedure and the implementation settings that complement the main method description.

\begin{algorithm}[ht]
\caption{ROVE: Iterative policy improvement}
\label{alg:rove_policy_improvement}
\begin{algorithmic}[1]
\Require Initial policy $\pi_{\theta_0}$, initial actor and critic data buffers $\mathcal{D}^{\mathrm{actor}}_0$ and $\mathcal{D}^{\mathrm{critic}}_0$, number of iterations $K$
\For{$k=0,\ldots,K-1$}
    \State Train critic $V_{\phi_k}$ on $\mathcal{D}^{critic}_k$ with the OVE objective
    \State Compute advantages for each transition segment:
    \[
    A_{\phi_k}(s_t,a_{t:t+H-1}) =
    \sum_{i=t}^{t+H-1}\gamma^{i-t}r_i
    + \gamma^H V_{\phi_k}(s_{t+H})
    - V_{\phi_k}(s_t)
    \]
    \State Compute threshold $\eta_k$ from advantage statistics
    \State Assign binary improvement label
    \[
    I_t = \mathbf{1}\{A_{\phi_k}(s_t,a_{t:t+H-1}) > \eta_k\}
    \]
    \State Update actor to obtain $\pi_{\theta_{k+1}}$ with advantage-conditioned policy extraction:
    \[
    \max_{\theta}\;
    \mathbb{E}_{(s_t,a_{t:t+H-1},I_t)\sim\mathcal{D}^{actor}_k}
    \left[
    \log \pi_{\theta}(a_{t:t+H-1}|s_t,I_t)
    \right]
    \]
    \State Deploy $\pi_{\theta_{k+1}}$ to collect new rollouts and human intervention trajectories
    \State Update data buffer with robot experience $\mathcal{D}^{actor}_{k+1}$
    \State Update data buffer with both robot and human experience $\mathcal{D}^{critic}_{k+1}$
\EndFor
\end{algorithmic}
\end{algorithm}

\textbf{Action chunk horizons.}
At inference, and during critic training and advantage computation, we use a chunk horizon $H=16$. Specifically, the policy executes the first $16$ steps of each predicted chunk before replanning; OVE bootstraps over $16$-step segments when constructing TD targets; and advantage labels are assigned to $(s_t, a_{t:t+H-1})$ pairs with $H=16$. We use a longer training chunk of $50$ to improve temporal consistency during supervised learning, while the shorter operational horizon stabilizes value bootstrapping and makes advantage estimation more responsive to local progress changes during deployment and intervention data.

\textbf{Model architecture.}
We use a separate state-value critic and a VLA policy. The value model is initialized from the VLAC~\citep{zhai2025vision} checkpoint as its VLM backbone: we extract intermediate representations from transformer layer $23$ (hidden dimension $2048$) and regress a scalar value with a lightweight transformer value head. The policy is initialized from Qwen3-VL-4B-Instruct~\citep{Yang2025Qwen3TR}, consumes the final-layer backbone features (hidden dimension $2560$), and outputs continuous action chunks with a flow-matching diffusion-transformer (DiT~\citep{Peebles2022ScalableDM}) action decoder. For both models, we freeze the pretrained vision-language model and train only the value head and action head.

\textbf{Critic training.}
The critic is trained with TD bootstrapping and an expectile value loss with coefficient $\tau=0.7$. Optimization uses $8000$ steps on $8$ GPUs with $140$GB RAM, per-device batch size $64$, and learning rate $1{\times}10^{-4}$ for first iteration, and $1{\times}10^{-5}$ for subsequent iterations. At iteration $0$, the critic initializes from the pretrained value checkpoint; thereafter it initializes from the previous iteration's critic. TD targets are normalized to $[-1,0]$ before expectile regression.

\textbf{Actor training.}
The actor is fine-tuned with advantage conditioning for $8000$ steps on $8$ GPUs with $140$GB RAM, per-device batch size $16$, and learning rate $1{\times}10^{-4}$ for first iteration, and $1{\times}10^{-5}$ for subsequent iterations. At iteration $0$, the actor starts from the pretrained VLA checkpoint; for $k>0$, it initializes from the policy checkpoint of iteration $k-1$.

\textbf{Data buffers.}
$\mathcal{D}^{actor}_k$ stores robot experience used for policy learning: teleoperated demonstrations, autonomous rollouts, and human intervention trajectories. $\mathcal{D}^{critic}_k$ extends this robot data with human experience videos for value learning. 

\textbf{State regularization.}
State regularization is enabled only during training and disabled at evaluation. In the actor action head, full-state dropout is applied per sample with probability $0.3$. Gaussian state noise is applied per sample with probability $0.4$ and base standard deviation $0.01$, with smaller scales for head and waist state blocks and unit scale for arm, hand, and end-effector state blocks. In the data transform, perturbation adds Gaussian noise to configured state/action keys and clips the result to the original value range; dropout zeros the configured keys when sampled. These settings are applied to proprioceptive inputs rather than image tokens.

\begin{table}[t]
    \centering
    \small
    \caption{Data scale used for iterative policy improvement. Counts report base-policy demonstration episodes and newly collected episodes at each iteration; intervention fraction denotes the fraction of episode time under human intervention, computed over rollout--intervention episodes.}
    \label{tab:data_scale_episodes}
    \renewcommand{\arraystretch}{1.12}
    \setlength{\tabcolsep}{5pt}
    \begin{tabular}{lccccc}
    \hline
    Task & Demo episodes & Iter.~1 & Iter.~2 & Iter.~3 & Intervention fraction \\
    \hline
    \textit{Erase the whiteboard} & 225 & 82 & 71 & 79 & 25.50\% \\
    \textit{Put the bread into the toaster} & 220 & 97 & 69 & 104 & 4.53\% \\
    \hline
    \end{tabular}
    \end{table}

\textbf{Data scale.}
\label{sec:data_scale}
Table~\ref{tab:data_scale_episodes} summarizes episode growth across iterations and intervention-time fractions. The base policies are trained only on teleoperated demonstrations. For iterative training, each round adds newly collected robot experience from policy deployment and human intervention.
For \textit{Erase the whiteboard}, interventions are typically triggered when the policy is about to fail in picking up the eraser, wipes in a direction that does not advance toward the handwriting, or leaves the whiteboard partially erased. For \textit{Put the bread into the toaster}, interventions mainly occur after the policy has made repeated insertion attempts 3 to 5 times but remains misaligned with the toaster slot. Since these autonomous alignment attempts can occupy a relatively long portion of the episode before takeover, the intervention fraction is lower than in \textit{Erase the whiteboard}.

Beyond robot data, we additionally collect $180$ ego-centric human experience videos per task using a head-mounted capture device, split evenly between successful and failed executions. These videos reproduce the behavior patterns observed in robot rollouts and interventions and further extend the coverage of partial-progress, near-failure, and recovery states, providing additional cross-embodiment supervision for value learning.

\section{Consecutive success examples}
\label{sec:success}

\begin{figure}[t]
    \centering
    \includegraphics[width=1.0\linewidth]{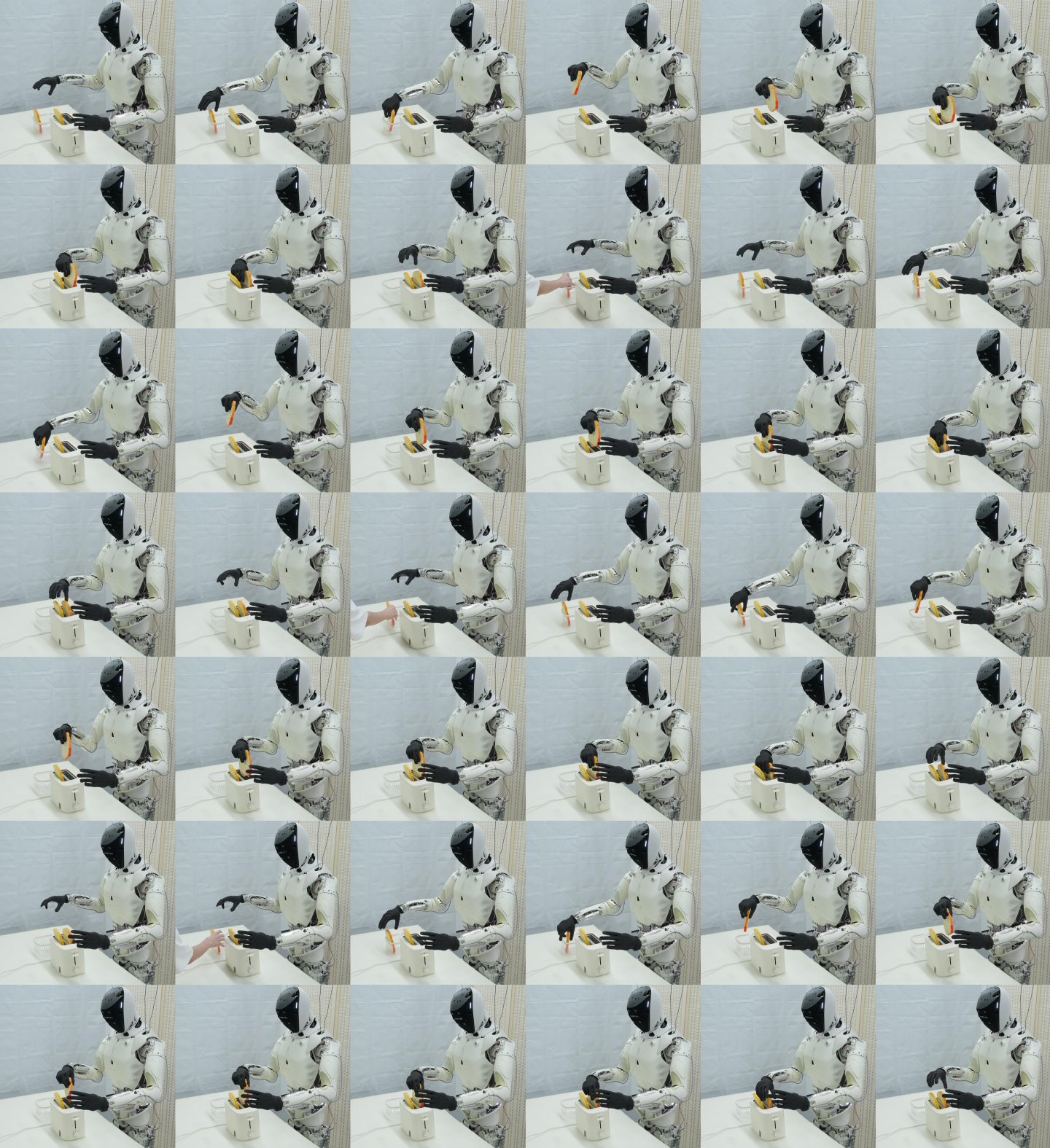}
    \caption{Four consecutive successful trials on \textit{Put the bread into the toaster}. Frames are arranged in a $7\times6$ grid and read left to right, top to bottom, in temporal order.}
    \label{fig:bread_continue_success}
\end{figure}

\begin{figure}[t]
    \centering
    \includegraphics[width=1.0\linewidth]{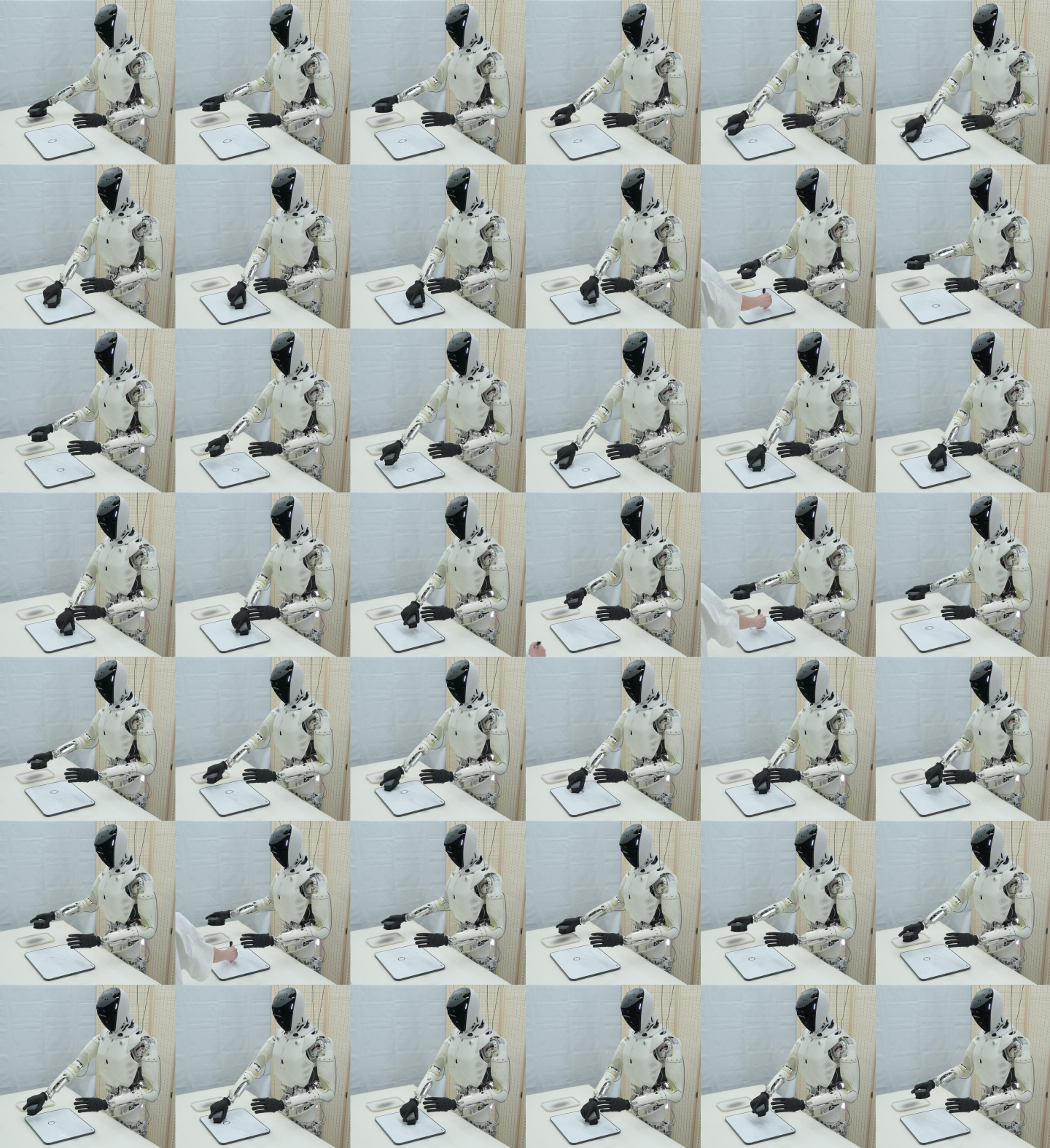}
    \caption{Four consecutive successful trials on \textit{Erase the whiteboard}. Frames are arranged in a $7\times6$ grid and read left to right, top to bottom, in temporal order.}
    \label{fig:erase_continue_success}
\end{figure}

Fig.~\ref{fig:bread_continue_success} and Fig.~\ref{fig:erase_continue_success} visualize four consecutive successful trials from our policy on \textit{Put the bread into the toaster} and \textit{Erase the whiteboard}, respectively.

\end{document}